\pdfoutput=1
\documentclass[11pt]{article}

\usepackage{acl}

\usepackage{times}
\usepackage{latexsym}

\usepackage[T1]{fontenc}
\usepackage[utf8]{inputenc}

\usepackage{microtype}

\usepackage{graphicx}

\title{Multi-hop Evidence Pursuit Meets the Web: Team Papelo at FEVER 2024}

\author{Christopher Malon \\
  NEC Laboratories America \\
  Princeton, NJ 08540 \\
  \texttt{malon@nec-labs.com} \\}

\newcommand{\abs}[1]{\vert#1\vert}

\begin{document}
\maketitle
\begin{abstract}
Separating disinformation from fact on the web has long challenged both
the search and the reasoning powers of humans.  We show that the reasoning
power of large language models (LLMs) and the retrieval power of modern
search engines can be combined to automate this process and explainably
verify claims.  We integrate LLMs and search under a {\em multi-hop
evidence pursuit} strategy.  This strategy generates an initial question based on
an input claim using a sequence to sequence model,
searches and formulates an answer to the question, and iteratively
generates follow-up questions to pursue
the evidence that is missing using an LLM.
We demonstrate our system
on the FEVER 2024 (AVeriTeC) shared task.  Compared to a strategy of
generating all the questions at once, our method obtains .045 higher label
accuracy and .155 higher AVeriTeC score (evaluating the adequacy of the
evidence).
Through ablations, we show the importance of various design choices,
such as the question generation method,
medium-sized context, reasoning with one document at a time,
adding metadata, paraphrasing, reducing the problem to two classes,
and reconsidering the final verdict.
% Finally, we compare the impact of less powerful open source models
% in this system.
Our submitted system achieves .510 AVeriTeC score on the dev set and
.477 AVeriTec score on the test set.
\end{abstract}

\section{Introduction}

Since 2018, the FEVER shared task has challenged natural language
processing systems to verify claims using a corpus and provide
evidence that witnesses these verdicts.
It has evolved from a simple
combination of natural language inference (NLI) and entailment \citep{thorne-etal-2018-fact}
to a challenge involving adversarially constructed claims \citep{thorne-etal-2019-fever2},
to a challenge to verify complex, multi-hop claims using a combination
of tables and free text \citep{aly-etal-2021-fact}.  In the current task, it
finally arrives at combating real-life disinformation on the web
\citep{averitec}.  

Systems are challenged to classify claim texts as supported, refuted,
not enough evidence, or conflicting evidence/cherrypicking.  In addition
to classifying the claim, the systems must submit a list of
questions and answers about a claim as evidence, with each answer derived
from information on the open web and cited with a URL. 
Credit is given only when both the classification matches the ground truth
and the evidence is adequate.  The AVeriTeC score determines evidence
adequacy by thresholding an average of METEOR scores between each gold QA
pair and the corresponding submitted QA pair in the best assignment of
QA pairs.

This task may involve retrieval and reasoning skills
at a level for which professional journalists are sometimes employed.
The reasoning may involve quote verification, stance detection, or
numerical comparisons.  The retrieval challenge goes beyond
previous political fact-checking
tasks \citep{politihop, alhindi-etal-2018-evidence}  % alhindi is Liar-Plus
and even beyond previous FEVER tasks in advancing from a closed corpus
(Wikipedia) to the open web.

Whereas previous FEVER shared tasks needed to be solved by researcher-trained
models, the current shared task allows the use of commercial API
components.  The winning team in FEVEROUS based their retriever on
fitting a Dense Passage Retriever \citep{karpukhin-etal-2020-dense} to the FEVEROUS data
\citep{bouziane-etal-2021-fabulous}, but the training data for FEVER 2024 is
quite limited, consisting of only 3,068 claims, and a retriever trained on
user feedback from worldwide search queries should easily be more powerful.
Additionally, an external web search engine such as Google Search may provide additional query understanding
features not found in DPR, as a recent feature (not in the API we used)
applies generative AI to search\footnote{\tt https://blog.google/products/search}.
Even though the gold evidence documents are guaranteed to appear in
the knowledge store provided by the contest organizers, the snippets may
not be extracted successfully.  We found that 297 of the 500 claims in the
dev set included gold documents with empty extracted text.  In contrast,
web search provides at least some text even from pages that the provided
web scraper is blocked from accessing.  Therefore, we chose to incorporate
web search into our system.

Relying on a large language model (LLM)
such as GPT-4o \citep{gpt4} for reasoning lets us leverage skills that
could not be learned from 3,068 heterogenous claims, and go beyond the
simple semantic comparison of an NLI model.  Beyond simple NLI,
ChatGPT and GPT-4 have been utilized to detect hallucinations in
text summaries \citep{chatgptsummaries}, as multi-faceted evaluators
that score generated text \citep{mtbench}, and for critiques and corrections
of generated text \citep{criticbench}.

Though there are many ways of using a search engine and LLM within
a fact-checking system, our main contribution is to show the power of combining 
them in a strategy of {\em multi-hop evidence pursuit}, which formulates
additional questions only after searching and formulating answers to
previous questions.  In the following sections, we also investigate
the impact of many choices of how the questions could be generated,
the nature and size of context for generating answers, handling of multiple
search results, metadata, paraphrasing, reducing the problem to two classes,
and reconsidering the final verdict.

We release a reference implementation of our system.\footnote{\tt github.com/cdmalon/fever2024}

\section{Related work}

Retrieval-augmented generation (RAG) \citep{rag} provides a general paradigm
for enabling an LLM to answer questions that surpass the knowledge encoded in
the LLM parameters, which is a task somewhat isomorphic to verifying claims
\citep{qa2d}.

A growing body of work utilizes LLMs as high-level reasoning controllers that
can solve tasks by querying agents to provide information or solve
subproblems \citep{agentsurvey, autogen}.
An early example for fact-checking an LLM's own output was
LLM-Augmenter \citep{checkyourfactsmsr}, which called an open retrieval
pipeline as an agent action to iteratively improve an LLM response.
\citet{rqrag} uses an LLM to rewrite, decompose, and disambiguate
queries before searching, and these steps are made into a hierarchy
of agents in \citet{mindsearch}.
\citet{factcheckgpt} used a combination of Google search and GPT-4 with
a single hop to fact-check claims in the FacTool-KB, FELM-WK, and HaluEval
datasets.
Behind a closed API, SearchGPT has been launched in beta to a few users
as a service to provide access to a search-empowered OpenAI
LLM.\footnote{\tt openai.com/index/searchgpt-prototype/}

FEVER 2024 presents a multi-hop, open corpus fact verification challenge.
In the multi-hop shared task of FEVEROUS, all but two
contestants collected all the needed evidence up front, after only
reading the claim \citep{aly-etal-2021-fact}.  Later top performers
(DCUF, UniFee, SEE-ST) addressed evidence interaction with graph-based methods
but still did not address evidence that might be missed by the initial document
retrieval
\citep{hu-etal-2022-dual, hu-etal-2023-unifee, wu-etal-2023-enhancing}.
\citet{malon-2021-team} established an iterative paradigm for fact verification
that retrieves further documents, sentences, and table cells by generating
follow-up queries that are formulated after considering only the first
retrieval, which we follow in the present system, in {\em multi-hop
evidence pursuit}.

In medical question answering, \citet{medfollowup} contemporaneously has
proposed ``iterative RAG for medicine'' which uses an LLM to generate
follow-up questions considering previous retrievals.  In our algorithm,
the relevance of each question is assured by generating it only upon
a failure to verify the claim as true or false based on the existing
evidence.  Their method may generate irrelevant questions after an answer
could already be obtained, simply because the fixed numbers of questions are
not achieved, resulting in lower evidence relevance and higher computational
cost.  Our system can stop as soon as a verdict is clear, and if our system
is configured to generate additional questions by paraphrasing, their
relevance is assured by their similarity to the original questions.

\section{Methodology}

\begin{figure*}[htb]
\begin{center}
\includegraphics[height=3in,width=6in]{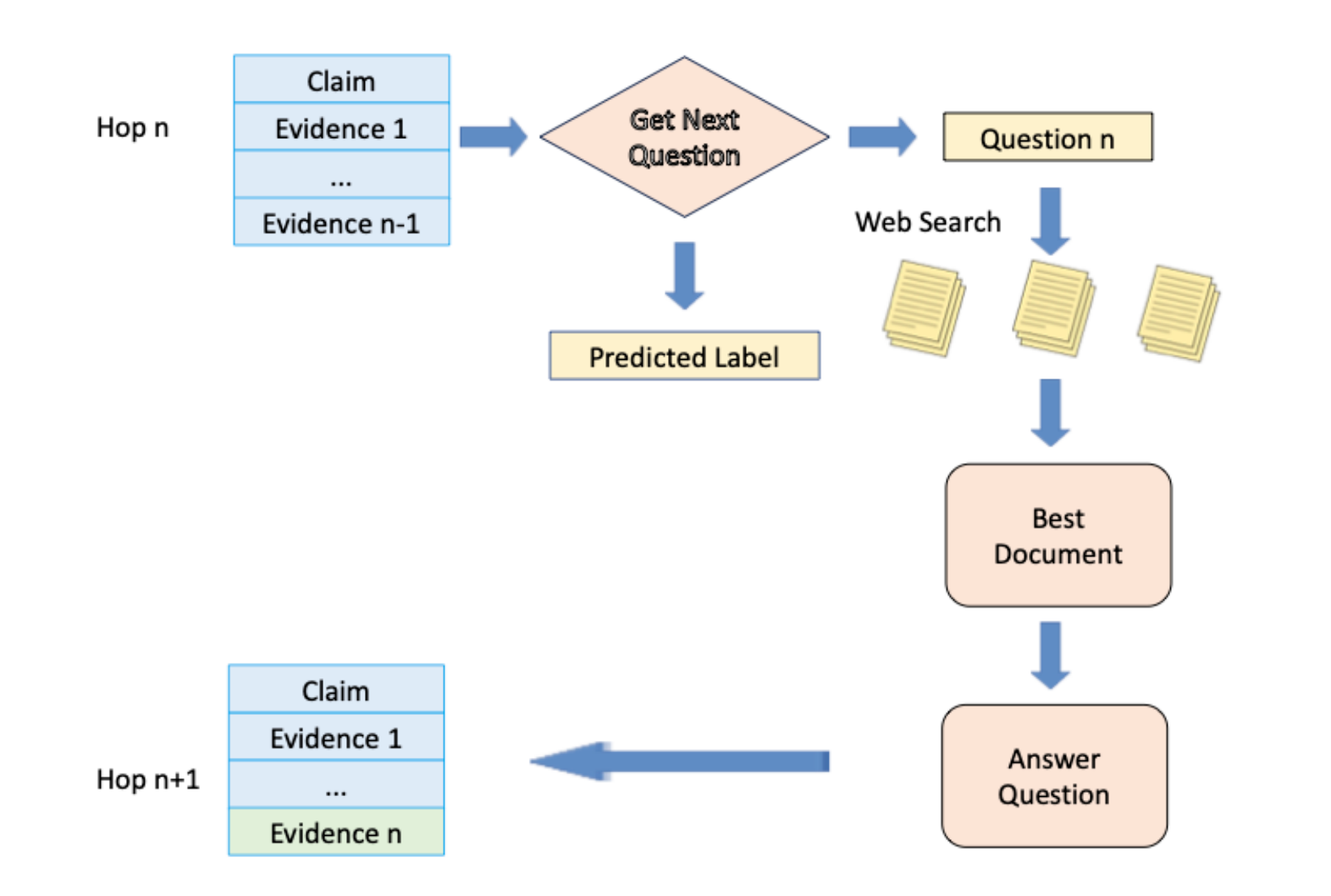}
\caption{Pursuing additional evidence by generating follow-up questions.}
\label{fig:nexthop}
\end{center}
\end{figure*}

\subsection{Overall architecture}

Pseudocode outlining the overall system is given in Algorithm 1,
with the main loop shown in Figure~\ref{fig:nexthop}.
At the core of the system are question generation functions
$GetFirstQuestion$ and $GetNextQuestion$, for which we consider
implementations either by sequence-to-sequence encoder-decoder
transformers such as T5 \citep{t5}, or by an LLM.
The $GetAnswer$ function (Algorithm 2)
prompts an LLM to implement $LLMBestDoc$ and $LLMAnswer$ to
answer the generated questions.  The final verdict is also
chosen by prompting an LLM with the generated questions and
answers, in $LLMVerdict$.

\begin{table}[htb]
\begin{center}
\begin{tabular}{l}
\hline
{\bf Algorithm 1.}  Claim verification \\
\hline
\textbf{Input:} Claim $c$, max questions $n$ \\
Initialize QA list $Q = \emptyset$ \\
Let $q = GetFirstQuestion(c)$ \\
\textbf{while} $\abs{Q} < n$ and $q \ne True$ and $q \ne False$ \\
\; Let $a = GetAnswer(q, c)$ \\
\; Append $(q,a)$ to $Q$ \\
\; Let $q = GetNextQuestion(c, Q)$ \\
\; {\em \# $GetNextQuestion$ outputs True or False} \\
\; {\em \# if next question not needed} \\
Let $k = \abs{Q}$ \\
\textbf{while} $\abs{Q} < n$ \\
\; Let $i = \abs{Q}$ \\
\; Let $q = Paraphrase(q_{i\,mod\,k})$ \\
\; Let $a = GetAnswer(q, c)$ \\
\; Append $(q,a)$ to $Q$ \\

\textbf{Output:} $v = LLMVerify(Q, c)$ and $Q$ \\
\hline
\end{tabular}
\label{tbl:algorithm}
\end{center}
\end{table}

\begin{table}[htb]
\begin{center}
\begin{tabular}{l}
\hline
{\bf Algorithm 2.} Function $GetAnswer(q,c)$ \\
\hline
\textbf{Input:} Question $q$, claim $c$ \\
Let $s = c + q$ concatenation \\
Let $G = WebSearch(s)$ \\
\textbf{if} $G = \emptyset$: \\
\; Let $G = WebSearch(NamedEntities(s))$ \\
$G = \{(url_0, quote_0), \ldots, (url_{9}, quote_{9})\}$ \\
Let $i = LLMBestDoc(G, q)$ \\
Let $d = FullDocument(url_i)$ \\
Let $e = AlignContext(d, quote_i, 5)$ \\
\textbf{Output:} $a = LLMAnswer(q, e)$ \\
\hline
\end{tabular}
\label{tbl:getanswer}
\end{center}
\end{table}

Unlike the baseline system \citep{averitec}, our system does not
generate questions on a {\em post hoc} basis after finding evidence,
but generates questions before web searches, playing a key role
in steering the verification process.  Rather than assuming
all evidence can be found up front with a single search query,
we review the current set of evidence and generate text
(in our case, a question) that provides a query to search for what is
still missing and needed after each hop, like the followup queries
introduced in \citet{malon-2021-team}.  Whereas the queries in
\citet{malon-2021-team} were generated by training a sequence to
sequence model to predict what the missing evidence would look like,
our system prompts an LLM to ask a question that the missing evidence answers.

The generation of evidence QA pairs temporarily stops when $GetNextQuestion$
thinks it can classify the claim as supported or refuted without
asking another followup question (see Appendix \ref{sec:prompts}).
After that point,
the already generated questions are paraphrased using an LLM and corresponding
answers are found until the desired number of QA pairs is obtained.
Finally, an LLM uses all QA pairs to decide the final classification
for the claim.

\subsection{Question generation}

\label{subsec:methodqg}

We consider two variants for the functions $GetFirstQuestion$ and
$GetNextQuestion$.  In the \textbf{Seq} version, we finetune
a sequence-to-sequence encoder-decoder transformer model.
For $GetFirstQuestion$, the input is the claim, and the output is
the first question.  For $GetNextQuestion$, the input is the claim
concatenated with all previous question-answer pairs, in the format
\begin{quotation}
Claim: {\em claim} Question: $question_0$ Answer: $answer_0$
Question: $question_1$ Answer: $answer_1 \ldots$
\end{quotation}
and the output is the next question to be generated.  These input strings
are prefixed with the string ``question: ''.
Details of the fine-tuning procedure are in Appendix \ref{sec:finetuning}.
Question-answer pairs from the gold data in the training set are used
for this fine-tuning.

The other variant is the \textbf{LLM} version, in which we
prompt the LLM with similar inputs.  The prompts are given in
Appendix \ref{sec:prompts}.  Because LLM output is often verbose
and may contain unnecessary explanations, we sentence split the output
and use only the first sentence containing a question mark.
If this is impossible, we use the whole output.

If an adequate number of questions and answers has been generated
and the verdict is clear, the model has the opportunity to output a
{\em True} or {\em False} verdict to stop the question generation.

As a further ablation, we consider a more traditional technique of
generating all the questions at once, given the claim.  The function % 007 cite
$AllAtOnce$ (prompt in Appendix \ref{sec:prompts}) replaces $GetFirstQuestion$
to generate a set of questions, and the \textbf{while} loop
in Algorithm 1 is replaced by a loop over the generated questions,
calling $GetAnswer$ but not $GetNextQuestion$.

\subsection{Evidence selection}

Here we describe the function $GetAnswer$, displayed in Algorithm 2,
which retrieves evidence and uses it to answer the generated questions.
Prompts for its LLM helper functions are given in Appendix \ref{sec:prompts}.

The generated question is concatenated to the claim to form a web search query,
and the top ten search results are obtained, including their URL,
the short snippet displayed in the search results, and usually the
page title, site name, and publication date.  When the web search
returns no results, we retry the search using only the named entities (and
other capitalized words after the first word) from the initial search query,
following the supplemental queries which improved retrieval by Wikipedia
page title lookups in \citet{malon-2018-team}.

By prompting, $LLMBestDoc$ is used to choose one document
that best answers the question
from the set of ten web search hits.  We attempt to retrieve and scrape
the text of that document using its URL (function $FullDocument$).
This is implemented using the \verb+scrape_text_from_url+ function
provided in the AVeriTeC baseline \citep{averitec}, which uses the
Python {\em trafilatura} library.\footnote{\tt github.com/adbar/trafilatura}
If the scraping succeeds,
we look for a small window of text (five sentences in our experiments)
that best overlaps the web search snippet (function $AlignContext$).
Specifically, all five-sentence windows of the document that include
more than 70\% of the words in the web search snippet are recorded in order,
and the middle such window is taken.
Using this window as the document excerpt provides more
background and context to the text that web search found to be relevant,
while avoiding prompting with the overwhelming amount of text that
might be found in the full web page.  If the scraping fails, we
continue to the next stage using only the web search snippet as document text.

Because $LLMBestDoc$ depends on parsing LLM output, it may fail to choose
a best document.
If a best document is chosen and the scraping succeeds, the LLM is prompted
to answer the question using the selected five-sentence window of
the best document in $LLMAnswer$.  If the best document is chosen and
the scraping fails, $LLMAnswer$ is run using the text of the web search snippet
only.  If a best document was not chosen in $LLMBestDoc$, we use the full
text of the LLM response in that function as the answer and the web search
result page itself as the evidence.

In $LLMBestDoc$ and $LLMAnswer$, the prompt includes not only the
text for each document, but metadata including the page title, site name, and
publication date, when this metadata appears in web search results.
This metadata may occasionally be useful in assessing the credibility
or relevance of the information to the question.

\subsection{Reconsideration and Classification}

The $Paraphrase$ function asks the LLM for paraphrases of the existing
questions.  In practice, multiple paraphrases of each question are
requested at once to avoid repeated calls, even though they are used
one at a time.  Although these paraphrases may not be logically
necessary once $GetNextQuestion$ has determined a verdict,
sometimes they provide a chance to reconsider the same
questions using multiple sources.  The variations in wording also improve
the AVeriTec score, as discussed in section \ref{sec:experiments}.

The $LLMVerdict$ function is called after all question-answer
pairs are collected, to choose the predicted label for each example.
Using additional QA pairs, it may override the decision that
stopped the QA generation process.  Table \ref{tbl:labels} shows the
distribution of labels in the training and development sets.
``Not Enough Evidence'' and ``Conflicting evidence / cherrypicking''
are minority classes, and we were unable to predict them with good F1
score.  We obtained a higher score by limiting $LLMVerdict$ to
predicting ``Supports'' or ``Refutes.''

\begin{table}[htb]
\begin{center}
\begin{tabular}{lcc}
\hline
Class & Train & Dev \\
\hline
Supported & 27.7\% & 24.4\% \\
Refuted & 56.8\% & 61.0\% \\
NEI & 9.2\% & 7.0\% \\
Conflicting & 6.4\% & 7.6\% \\
\hline
\end{tabular}
\caption{Distribution of class labels.}
\label{tbl:labels}
\end{center}
\end{table}

\section{Experiments}
\label{sec:experiments}

\begin{table*}[htb]
\begin{center}
\begin{tabular}{lcccccc}
\hline
System & Supp F1 & Ref F1 & NEI F1 & Conf F1 & Acc & AVeriTec 0.25 \\
\hline
$AllAtOnce$ & .591 & .813 & 0 & 0 & .705 & .340 \\
LLM+LLM  & .644 & .821 & 0 & 0 & .720 & .385 \\
Seq+Seq & .638 & .816 & 0 & 0 & .715 & .370 \\
4 class & .486 & .593 & {\bf .148} & {\bf .069} & .415 & .245 \\
No late verdict & .643 & .811 & 0 & 0 & .705 & .450 \\
No long doc & .577 & .819 & 0 & 0 & .705 & .465 \\
Multi-doc & .673 & .837 & 0 & 0 & .735 & .460 \\
No metadata & .575 & .810 & 0 & 0 & .700 & .410 \\
No paraphrase & .701 & .839 & 0 & 0 & .745 & .225 \\
Repeat not para & .624 & .813 & 0 & 0 & .710 & .340 \\
Algorithm 1 & {\bf .716} & {\bf .841} & 0 & 0 & {\bf .750} & {\bf .495} \\
\hline
\end{tabular}
\end{center}
\caption{Results on the first 200 examples of the dev set}
\label{tbl:results}
\end{table*}

\begin{table*}[htb]
\begin{center}
\begin{tabular}{llcccccc}
\hline
Data & Submission & Supp F1 & Ref F1 & NEI F1 & Conf F1 & Acc & AVeriTec 0.25 \\
\hline
Dev & Algorithm 1 & .698 & .853 & 0 & 0 & .754 & .486 \\
Dev & Inflated to 10 & .698 & .853 & 0 & 0 & .754 & .510 \\
Test & Algorithm 1 & --- & --- & --- & --- & --- & .445 \\
Test & Inflated to 10 & --- & --- & --- & --- & --- & .477 \\
\hline
\end{tabular}
\end{center}
\caption{Final results on full datasets}
\label{tbl:final}
\end{table*}

We implement Algorithm 1 using GPT-4o (\verb+gpt-4o-2024-05-13+, seed 42)
as the LLM, T5 (\verb+t5-large+) \citep{t5} as the sequence-to-sequence model,
and Google as the web search engine, and consider various ablations.
For a faster development cycle and reduced monetary cost,
Table~\ref{tbl:results} reports the performance of each of our systems
only on the first 200 examples of the development set.

\subsection{Question formation}

Recall from Section~\ref{subsec:methodqg} that in Algorithm 1,
the functions $GetFirstQuestion$
and $GetNextQuestion$ could be implemented either by \textbf{Seq} or
\textbf{LLM}, or instead of Algorithm 1, the questions could be generated
$AllAtOnce$.
Whichever question generation approach is used, at most five questions are
taken from the question generator and the paraphrase loop of Algorithm 1
extends the list to five questions.
The submitted system follows Algorithm 1 using \textbf{Seq}
for $GetFirstQuestion$, and \textbf{LLM} for $GetNextQuestion$ (Seq+LLM).

The lower performance of the $AllAtOnce$ alternative indicates that
this task requires followup searches considering the evidence already
retrieved, with searches that cannot be anticipated using the claim alone.
It validates our choice to use a {\em multi-hop evidence pursuit} strategy
\citep{malon-2021-team}.

The LLM+LLM
alternative shows that performance worsens if we generate the
first question using GPT-4o.  An inspection of the data
revealed that the gold first questions were usually simple rephrasings of
the claims, which T5 can learn well, whereas GPT-4o often tried to
generate something more complicated.

The Seq+Seq alternative shows that performance worsens if we generate the
subsequent questions using T5.  Subsequent gold questions often
reflected deeper reasoning using the obtained answers, which we suspect
are beyond the capabilities of simple sequence to sequence models.

\subsection{Label prediction}

We have implementations of $LLMVerdict$ that use a four-class prompt,
or eliminate the ``Not Enough Evidence'' (NEI) and ``Conflicting Evidence /
Cherrypicking'' classes to decide only between ``Supported'' and
``Refuted.''
The 4-class result (otherwise the same as the main system) shows very low
F1 scores for the NEI and Conflicting classes.  As NEI claims form
only 7.0\% of the dev set and Conflicting claims form only 7.6\%, we decided
that it is always best to guess another label.

Another variant, ``No late verdict,'' calls $LLMVerdict$ only if the
\textbf{while} loop is not terminated by predicting True or False,
and maintains that early decision even after the paraphrases are added.
(If True is obtained, ``Supported'' is predicted and if False is obtained,
``Refuted'' is predicted.)  The difference in label accuracy shows it is
sometimes useful to consider the whole question and answer chain from the
beginning when forming a verdict.

\subsection{Answer formation}

The submitted system uses $FullDocument$ and $AlignContext$ to obtain
longer document contexts for prompting $LLMAnswer$.
The ``No long doc'' ablation uses only the original web search snippet
as context for $LLMAnswer$.  The close performance in
AVeriTeC score shows that while longer context is helpful, it is often
unnecessary.  Scraping web pages to obtain this longer context has become
difficult as many sites seek to restrain robots, so relying on snippets
is convenient.  In cases where our scraping fails, the original
snippet is returned by $FullDocument$ anyway.

The ``Multi-doc'' ablation calls $LLMAnswer$ using all ten search hits
and their snippets, without calling $LLMBestDoc$ to focus on one.
It is very close to our system in label accuracy.  Although it narrows the
depth and context of information presented to $LLMAnswer$, it may have
advantages in presenting multiple possible perspectives.

Metadata for each document context is usually presented to $LLMAnswer$
in the form $$\textrm{Document\,} i: (title, \textrm{from\,} site,
\textrm{published\,} date)$$
The lower label accuracy and AVeriTeC score of the ``No metadata'' variant
show that knowing where evidence came from is helpful to the LLM.

\subsection{Evidence length}

When the label is predicted correctly for an example, the AVeriTeC score
thresholds an example score, which is computed as the sum of the METEOR scores
between gold QA pairs and best matching predicted QA pairs, divided by
the number of gold QA pairs.  Whenever fewer QA pairs are predicted
than gold QA pairs, those gold QA pairs contribute zero to this average.
Therefore, to optimize the AVeriTec score, it is important to predict at least
as many QA pairs as the number of gold pairs, even if the some predicted pairs
match poorly.

A system could submit up to ten QA pairs for each example.
However, only 5\% of examples had more than five gold QA pairs
in the development set.  Since the ultimate objective is optimizing
human evaluation rather than AVeriTeC score and reading more than five
QA pairs may be frustrating for a human, we initially applied our systems
to produce five QA pairs per question.

For many examples, Algorithm 1 could reach decisions of $q=True$ or $q=False$
in its first loop of $GetFirstQuestion$ and $GetNextQuestion$
using fewer than five QA pairs.
We compared the score obtained by
{\em repeating} QA pairs, or by asking the LLM to {\em paraphrase} the
existing questions in the second loop of Algorithm 1, until five QA pairs
were obtained.  In the case
of {\em paraphrase}, new answers are sought for the rewritten questions.
Besides improving the AVeriTeC score, the new answers may be used to
reconsider the final verdict.

The ``No paraphrase'' ablation has a minimal effect on label accuracy,
but since fewer QA pairs are generated, AVeriTec score is less than half
the score of the submitted system.  ``Repeat not paraphrase'' to get
five QA pairs can recover some of the AVeriTeC score, but the paraphrases
help the METEOR score of the best assignment much more than duplicates.

Ten QA pairs is the upper limit, and submitting additional QA pairs up to ten
can only improve the score of the best assignment between submitted pairs and
gold pairs.  We took our five generated QA pairs from Algorithm 1
($GetFirstQuestion$, $GetNextQuestion$, and paraphrasing) and
duplicated them to submit ten.  Naturally, repeating can be helpful if one
generated QA pair addresses points raised in multiple gold QA pairs.
The effect of inflating the QA pairs on our
full dev set and test set performance is shown in Table~\ref{tbl:final}.

\section{Conclusion}

The AVeriTeC shared task is a realistic fact-checking challenge on
actual web disinformation.  The best large language models offer
the deep reasoning power needed to pursue missing evidence to verify
claims, and the best web search engines provide the vast document indices
and retrieval capabilities needed to find it.

We have contributed a multi-hop evidence pursuit framework which
combines the strengths of sequence to sequence models with LLMs to
generate first question and subsequent questions separately, considering
the present information; to stop pursuit once the answer is clear; and to
embellish evidence by paraphrasing before considering the whole evidence
chain to make the final verdict.  Ablations indicate the importance of
each design choice.  Multi-hop evidence pursuit outperforms trying to
generate all questions in one step.  Reducing the number of classes,
and using metadata and multi-sentence context from one best document,
were important in obtaining our best performance.

The fact checking system presented may be useful to expedite the work
of human fact checkers or provide a more rapid preliminary response
to disinformation.  Its full explainability could mitigate the effect of
misclassifications, if the explanations were read and considered by
a human.  Over a history of many claims,
ratings of disinformation from our system and/or human fact checkers
could be used to rate the credibility of an information source.

\section*{Limitations}

When ``Not Enough Evidence'' (NEI) is an option, an LLM tends to select it too
often.  Our system was unable to predict either NEI or ``Conflicting
Evidence / Cherrypicking'' with acceptable accuracy.
Considering this, and the fact the overall label accuracy is only .754,
humans should be cautious in trusting this system's output
to verify a claim without reading the rationale.

LLMs have insufficient information to judge the overall credibility of
a website, and currently just the site name is given for the LLM's
consideration.  Metadata including the site name helps (to give an example
from the dev set, GPT-4o was aware or discovered through its searches that
Scoopertino was a satirical website), but generally, misinformation that is
corroborated elsewhere on the web may fool our fact checking system.

Although the LLM is always prompted to answer questions ``based on
the above information'' quoted from retrieved documents or its previous
answers, there is no guarantee that the LLM does not apply other,
untraceable knowledge in forming its answers.  We use a date filter
to ensure that all web searches return documents only from before each claim
date, but we use an LLM whose training cutoff is after the claim dates.

Novel information first reported, which has no basis in existing documents,
can never be fact-checked with the techniques of this system (for example,
the first report that a presidential candidate was shot).  That kind
of fact checking requires judgments of plausibility, credibility, and
consistency that are out of scope for this system.

\bibliography{anthology,fever2024}

\appendix

\section{Fine-tuning}
\label{sec:finetuning}

A \verb+t5-large+ model was fine-tuned for three epochs with batch size 4,
maximum source length 64 or 256 for $GetFirstQuestion$ or $GetNextQuestion$,
and maximum target length 64.  For the AdamW
optimizer, default Huggingface values of $5 \times 10^{-5}$ were used
for the learning rate, $\beta_1 = 0.9$, and $\beta_2 = 0.999$.
The model was prompted with the prefix ``question: '' followed by the inputs.
Only gold data from AVeriTeC was used for the fine-tuning of each model.

\section{Prompts}
\label{sec:prompts}

\textbf{GetFirstQuestion.}  For the LLM variant, the prompt is:
\begin{quotation}
We are trying to verify the following claim
by {\em speaker} on {\em date}.  Claim: {\em claim}  We aren't sure whether
this claim is true or false.  Please write one or more questions that would
help us verify this claim, as a JSON list of strings.  Keep the list short.
\end{quotation}
The JSON is parsed and only the first string in the list is used.

\textbf{AllAtOnce.}  For the $AllAtOnce$ variant, we use the same prompt
as $GetFirstQuestion$ to get the questions, but we keep the entire list.

\textbf{GetNextQuestion.}  For the LLM variant, the prompt is:
\begin{quotation}
We are trying to verify the following claim
by {\em speaker} on {\em date}.  Claim: {\em claim}  So far we have asked
the questions: Question 0: $question_0$ Answer: $answer_0$ Question 1:
$question_1$ Answer: $answer_1 \ldots$
Based on these questions and answers, can you verify whether the claim is
true or false?  Please answer \verb+[[True]]+ or \verb+[[False]]+, or ask
one more question that would help you verify.
\end{quotation}
The response is searched for \verb+[[True]]+ or \verb+[[False]]+.  If neither
is found, then the response is sentence tokenized with
the \verb+sent_tokenize+ function of NLTK 3.8.1 and the first sentence
that includes a question mark is returned.

\textbf{LLMBestDoc.}  The prompt is:
\begin{quotation}
We searched the web and found the following information.
Document 0 ($title_0$, from $site_0$, published $date_0$): $snippet_0$
Document 1 ($title_1$, from $site_1$, published $date_1$): $snippet_1$
\ldots
Document 9 ($title_9$, from $site_9$, published $date_9$): $snippet_9$
Based on the above information, please answer the following question,
referring to the one document that best answers the question.
$question$
\end{quotation}
Note that the original claim is not used in this prompt.  The response is
searched with a regex for the first instance of \verb/Document\s+([0-9])//
or \verb/Documents[ 0-9,]+and ([0-9]+)/ and the corresponding numbered
document is taken.  If the regex search fails, the search result page
itself is used as context for answering the question, and the full response
is used as the answer.

\textbf{LLMAnswer.}  Unlike $LLMBestDoc$, this is called with context
from one document.  The prompt is:
\begin{quotation}
We searched the web and found the following information.
Document ($title$, from $site$, published $date$): $context$
Based on the above information, please answer the following question.
$question$
\end{quotation}
The entire response is used as the answer.

\textbf{Paraphrase.}
The prompt is:
\begin{quotation}
Please give four ways to rephrase the following question.
Give your answer as a JSON list of strings, each string being one question.
Question: $question$
\end{quotation}

\textbf{LLMVerify.}
The prompt is:
\begin{quotation}
We are trying to verify the following claim: $claim$
Based on our web searches, we resolved the following questions.
$0$.  $question_0$ \; $answer_0$
\ldots
$k$.  $question_k$ \; $answer_k$
Is the claim (A) fully supported by the evidence, or (B) contradicted by
the evidence?  Please respond in the format \verb+[[A]]+ or \verb+[[B]]+.
\end{quotation}
We search the response for \verb+[[A]]+ or \verb+[[B]]+.  For the
four class variant, the end of the prompt is:
\begin{quotation}
Is the claim (A) fully supported by the evidence, (B) contradicted by
the evidence, (C) insufficient information, or (D) conflicting evidence?
Please respond in the format \verb+[[A]]+, \verb+[[B]]+, \verb+[[C]]+, or
\verb+[[D]]+.
\end{quotation}

\end{document}